\documentclass[lettersize,journal]{IEEEtran}
\usepackage{amsmath,amsfonts}
\usepackage{algorithmic}
\usepackage{algorithm}
\usepackage{array}
\usepackage{amsmath}
\usepackage[caption=false,font=normalsize,labelfont=sf,textfont=sf]{subfig}
\usepackage{textcomp}
\usepackage{stfloats}
\usepackage{url}
\usepackage{verbatim}
\usepackage{graphicx}
\usepackage{cite}
\usepackage{color, soul}
\usepackage{authblk}
\usepackage{hyperref}
\usepackage{listings}
\usepackage{booktabs}

\hypersetup{
    colorlinks=true,
    linkcolor=blue,
    filecolor=blue,
    urlcolor=blue,
    citecolor=green,
}
\begin{document}

\title{SAM-helps-Shadow:When Segment Anything Model meet shadow removal}

\author{Xiaofeng Zhang, ChaoChen Gu*, ShanYing Zhu

\thanks{Xiao Feng Zhang, Chao Chen Gu and Shan Ying Zhu are with Center for Intelligent Wireless Network and Collaborative Control, Shanghai Jiao Tong University, Shanghai, China. (email: framebreak@stju.edu.cn; jacygu@sjtu.edu.cn; shyzhu@sjtu.edu.cn }% <-this % stops a space
% \thanks{Hao Tang is with the Department of Information Technology and Electrical Engineering, ETH Zurich, Zurich 8092, Switzerland. email: hao.tang@vision.ee.ethz.ch}
\thanks{Manuscript received August 18, 2022}}

\markboth{}%
{Shell \MakeLowercase{\textit{et al.}}: A Sample Article Using IEEEtran.cls for IEEE Journals}


\maketitle

\begin{abstract}
The challenges surrounding the application of image shadow removal to real-world images and not just constrained datasets like ISTD/SRD have highlighted an urgent need for zero-shot learning in this field. In this study, we innovatively adapted the SAM (Segment anything model) for shadow removal by introducing SAM-helps-Shadow, effectively integrating shadow detection and removal into a single stage. Our approach utilized the model's detection results as a potent prior for facilitating shadow detection, followed by shadow removal using a second-order deep unfolding network. The source code of SAM-helps-Shadow can be obtained from {\textcolor[rgb]{0,0,1}{https://github.com/zhangbaijin/SAM-helps-Shadow}}

\end{abstract}

\begin{figure*}[h]
\centering
\includegraphics[width=5.5in]{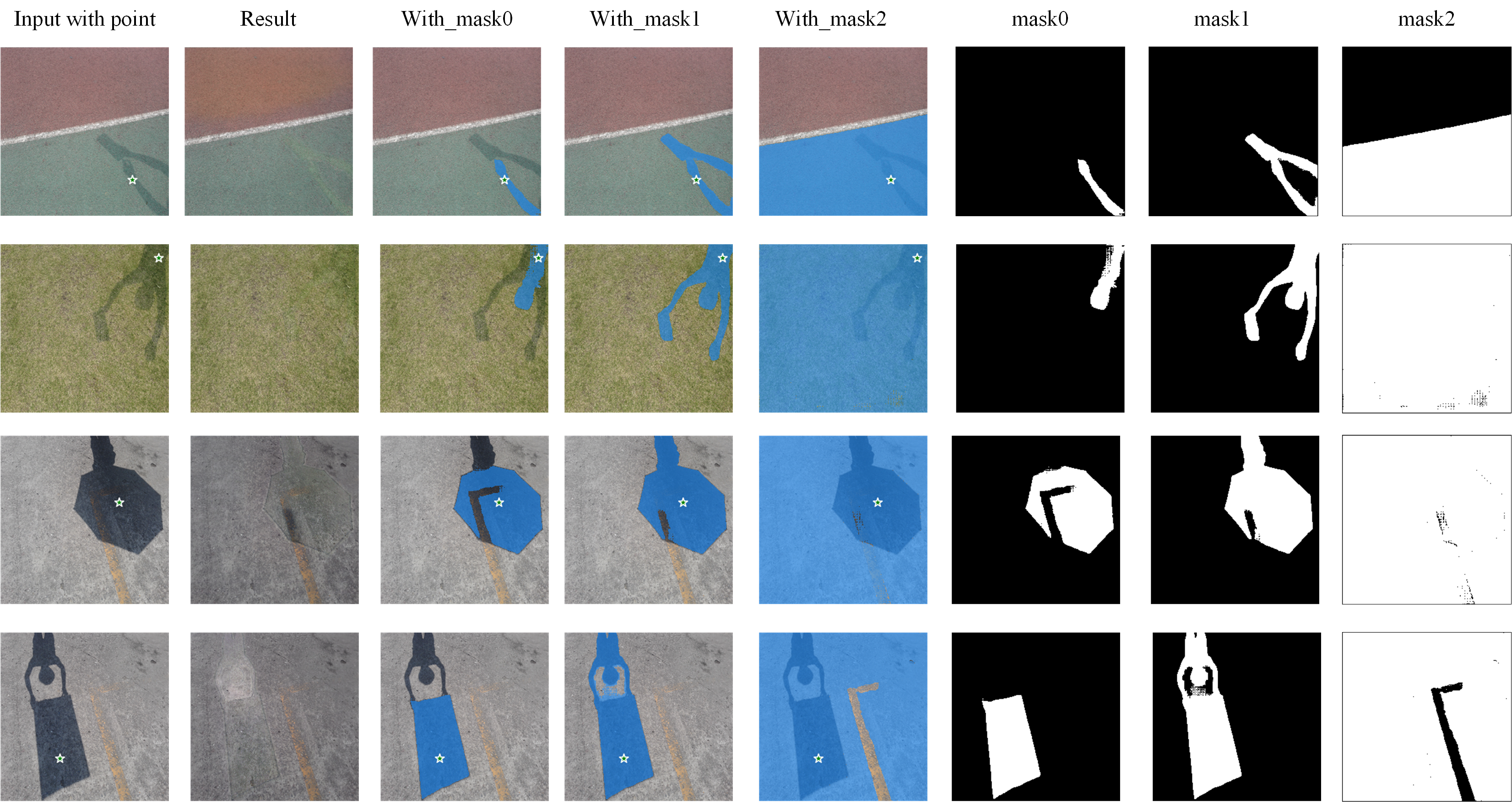}
\caption{The process of SAM-shadow(from left to right reprensent input/results/sam-shadow/mask)}
\label{introduction}
\end{figure*}

\section{Introduction}
The existence of shadows in natural images can provide information about the environment and illumination conditions, which in turn us to interpret the situation in the image, but also makes image processing more complex. Therefore shadows are often eliminated in the first stage of image processing, and the critical factor affecting the quality of elimination is the detection and localization of them.

%Deep learning still has issues with shadow identification, such as ambiguous shadow boundaries and rough features; as a result, the recovered shadow-free image after shadow removal using deep learning often includes artefacts and illumination condition changes. To address the aforementioned issues, a robust shadow detection and removal approach is urgently required.
Compared to the traditional image processing techniques \cite{lu2019,lu2017}, neural network \cite{lu2020,lu2018} has powerful representation ability. Recently, there are some works on deep-learning based shadow removal, such as self-supervised models based on GAN \cite{1,2,3,4,5,6,7,8}, semi-supervised models with mask-assisted guidance \cite{9,10,zhu2020,Jiang2022}, and some unsupervised learning methods \cite{11,12,13,14,22,23,24}.
\par
The biggest problem of image shadow removal is that it cannot be applied to actual images, and can only be limited to data like ISTD/SRD, so we urgently need zero-shot learning of image shadow removal. Currently, Segment anything model (SAM) \cite{inpainting-anything} has achieved unprecedented performance in natural image segmentation, while it considers shadows as background. Firstly, to make SAM suitable for shadow removal, we propose a method named ShadowNet-SAM, which combine shadow detection and shadow removal into one stage. Specifically, we employ its detection result as a strong prior to facilitate detecting shadows, Secondly, We remove the shadows using a second-order deep unfolding network. The result of ShadowNet-SAM as shown in Fig. \ref{introduction}. Secondly, the difficulty with shadow removal is that the nature of shadows is inherently complex. Not only do shadows change the brightness of an image, but they also change image properties such as color and texture. In addition, the shape and intensity of shadows change depending on the complexity of the light source and the scene. Our approach provides an effective solution to these challenges through zero-shot learning by SAM and deep unfolding network techniques.
The main advantage of using zero-shot learning techniques is that instead of relying on a large number of labeled samples, this approach segments the shadows by understanding the shadow properties in the image, alleviating the reliance on large amounts of labeled data and also greatly reducing the complexity of model training. In addition, the deep unfolding network is able to better handle the complex task of shadow removal by deconstructing layers, bringing higher accuracy and efficiency to shadow removal.
In summary, our method has three contributions: 
%This is because, in the CNN model, the number of operations required to compute the connection of two points using convolution increases with increasing distance between them. Therefore, it remains valuable and challenging to effectively remove shadows.
\par

\begin{enumerate}
 \item This paper is the first combine Segment anything model and  shadow removal into one stage, named SAM-helps-shadow.
 \item Instead of relying on a large number of labeled samples, this method of SAM detection of shadows segments shadows by understanding the shadow properties in images, alleviating the reliance on large amounts of labeled data and also greatly reducing the complexity of model training.
\item The experimental results on ISTD and SRD datasets prove that the proposed method achieves better performance and proves the effectiveness of the network.
\end{enumerate}

\section{Realted work}

\subsection{Inpaint Anything}

The application potential of the inpaint Anything (IA) \cite{inpainting-anything} in inpainting and related applications. IA (Input Anything) is a powerful and user-friendly pipeline that solves issues related to repair, such as object removal, content filling, and background replacement. It combines the advantages of segmenting any model (SAM), state-of-the-art image painters, and AI generated content (AIGC) models. IA provides three main functions:

1. Delete any content: Users can click on an object, and IA will delete it and fill in the "hole" with the surrounding context. 2. Fill anything: After the object is removed, the user can provide text based prompts, and IA will use AIGC models (such as stable diffusion) to fill the void with relevant generated content. 3. Replace anything: Users can choose to keep the selected object and replace the remaining background with the newly generated scene.

IA has demonstrated its potential in various repair scenarios, demonstrating its universality, robustness, and effectiveness. It can work at different content, resolutions, and aspect ratios. In addition, by revealing the potential of "composable AI", IA encourages the sharing and promotion of new projects based on its foundation. In the future, developers will continue to explore the functionality of IA, including its application in image shadow removal tasks. The goal is to improve its performance and versatility to address more repair related challenges.
\subsection{Segment anything model(SAM)}
Segment Anything Model (SAM) \cite{SAM} is a basic model that can handle various types of image segmentation tasks without retraining and has been evaluated to exhibit impressive zero-sample performance. 
\cite{medsam} and \cite{cmedsam} investigate the application of SAM to medical images. \cite{medsam} curates a large-scale medical image dataset and develops a simple fine-tuning method to adapt SAM to general medical image segmentation. \cite{cmedsam} uses a low-rank-based (LoRA) fine-tuning strategy to fine-tune the SAM image encoder with a cue encoder and a mask decoder for fine-tuning on a labeled medical image segmentation dataset. \cite{sammd} evaluates the potential of SAM for medical image segmentation tasks, and the results show that the model has a powerful zero-sample learning capability and generalizes well to abdominal CT data. \cite{rssam} develops an efficient pipeline for generating a large-scale Remote Sensing (RS) image segmentation dataset called SAMRS, using the SAM and existing RS object detection datasets. SAMRS surpasses existing high-resolution RS segmentation datasets in size by several orders of magnitude and provides object category, location, and instance information that can be used for semantic segmentation, instance segmentation, and object detection. The study provides a comprehensive analysis of SAMRS from various aspects and aims to facilitate research in RS image segmentation, especially in large model pre-training. \cite{samadpter} proposes SAM-Adapter to significantly improve the performance of SAM by incorporating domain-specific information or visual cues into the segmentation network. The method performs state-of-the-art tasks such as camouflaged object detection, shadow detection, and medical image segmentation. It opens up opportunities to apply SAM to downstream tasks in these domains. A new paradigm, "Segment Non-Euclidean Anything," is proposed in \cite{samdomain} to develop a basic model for processing diverse graphical data in non-Euclidean domains, and some solutions and future development directions are given. 

\section{Methodology}

We propose a new image shadow removal method that combines SAM and CNN. The following is the detailed content of the method chapter. (1) SAM contains a wealth of knowledge that can be transferred to any segmentation task  
(2) The fine-grained segmentation results provided by SAM can be used as the first stage of pre-processing for the shadow removal task, with the expectation that subsequent modules can be optimized to obtain better results  
(3) The prior knowledge related to the shadow removal task can be introduced through a careful design of the prompt to improve the final result and the interpretability of the method.

\begin{figure*}[h]
\centering
\includegraphics[width=5.5in]{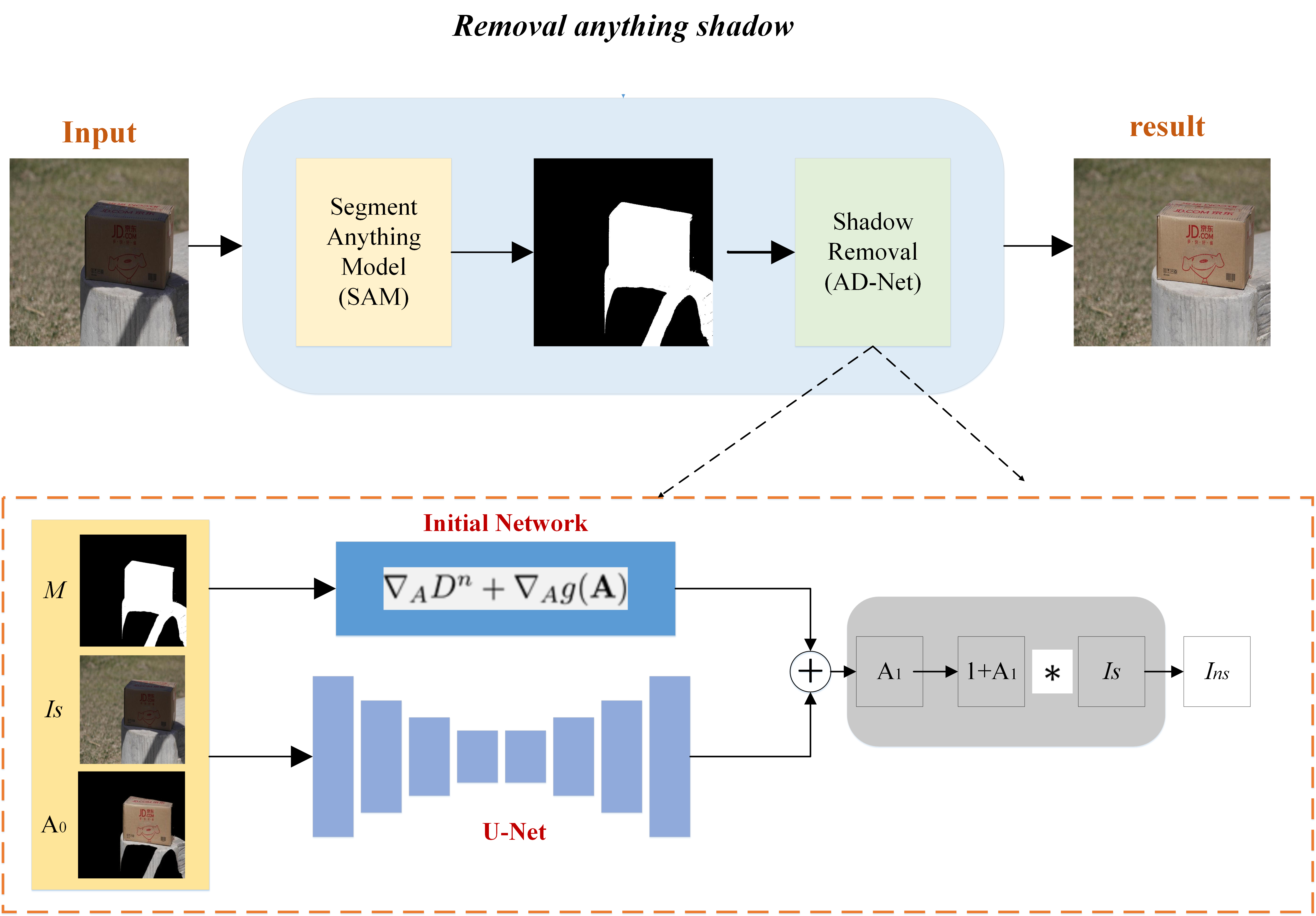}
\caption{The structure of SAM-shadow}
\label{structure}
\end{figure*}

\begin{figure*}[h]
\centering
\includegraphics[width=5in]{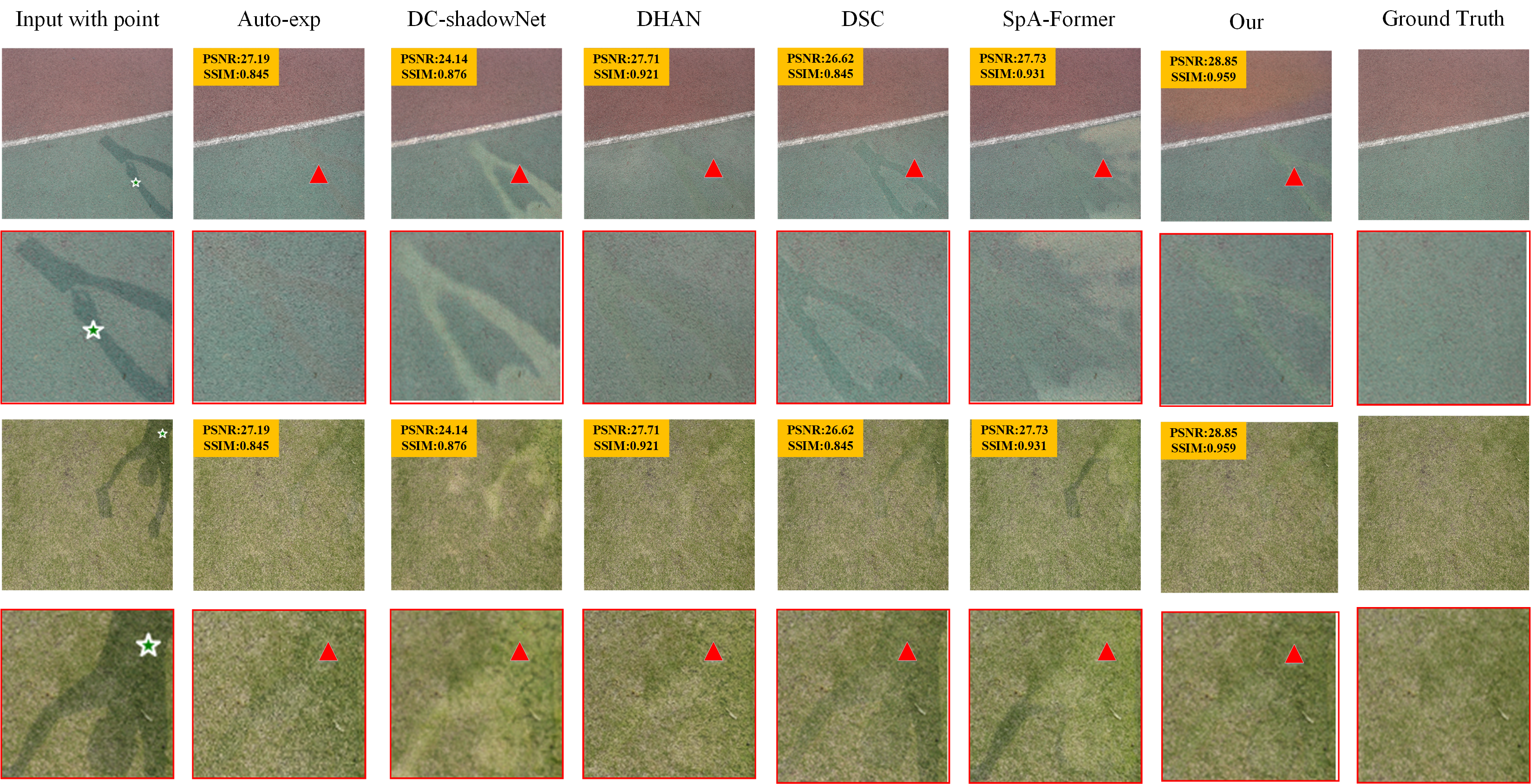}
\caption{The process of SAM-shadow(from left to right reprensent input/results/sam-shadow/mask)}
\label{compare}
\end{figure*}

\subsection{Shadow removal lighting model}

In order to effectively remove shadows from images while maintaining their spatial detail features, we propose a shadow removal illumination model. In this model, we first use a segmentation model based on the Segment Arbitrary Model (SAM) to accurately locate shadow areas. Then, we expand it using spatial variation characteristics and existing shadow masking information to improve the accuracy of the target mask. Specifically, we propose a new shadow illumination model that considers the spatial variation characteristics of the shadow region and utilizes existing shadow masking information. The new model can be represented as:
\begin{equation}
    S(x) = (1 + As) \times Is  \times Ms + (1 - Ms) \times Is
\end{equation}

S(x) is the image after shadow removal, As is the shadow removal feature parameter, and Ms is the target mask expanded based on the shadow region masking information.

We combine the second-order light enhancement curve IN2(x) with the shadow illumination model S(x) to further improve image quality. In this process, we simultaneously apply the light enhancement algorithm and shadow removal algorithm to achieve higher visual quality in the entire image:
\begin{equation}
     \text{Final}(x) = \alpha \times \text{IN2}(x) + (1 - \alpha) \times \text{S}(x) 
\end{equation}
Final(x) is the final processed imagze, and $ \alpha $ is a weight parameter used to balance the contributzions from the second-order light enhancement curve and shadow illumination model. By adjusting the value of $ \alpha $, we can achieve a smooth transition from just performing shadow removal ($ \alpha $=0) to just applying the second-order light enhancement curve ($ \alpha $ =1).

We introduced a shadow removal illumination model that combines the use of the Segment Arbitrary Model, spatial variation characteristics, and existing shadow masking information, and combined it with the second-order light enhancement curve to optimize low-light images. By adjusting the feature parameters A1, A2, As, and mask M, M2, and weight parameter $\alpha$, we can achieve high-quality output that removes shadows and enhances image brightness while preserving spatial detail features. Such an integrated approach is highly effective in improving image quality captured in shadow environments. 
\subsection{Variational Model of Iterative Algorithm CNN Framework}

We have designed a deep convolutional neural network based on iterative algorithms, combined with the U-net architecture and utilizing advanced deep learning technologies such as ViT and AIGC Models. At each iteration stage, we first update A, and then update the unshaded image Ins. Specifically, this process can be expressed as the following equations:

\begin{equation}
    A0, Ins =\operatorname{Ninit}(Is, M)
\end{equation}
\begin{equation}
    N_{a n+1}=\nabla A \phi\left(A_n\right)
\end{equation}
Where Is represents a shadow image, and the input image contains shadows. M represents a mask, identifying a binary image of shaded areas in the input image. A represents ambient light estimation, which is the intensity of light in the shaded areas of the input image. Ins represents a shadow removed image and outputs the resulting image after removing shadows.

$N_{an+1}$ and $Ninit$: represent different iteration stages, $Ninit$ is the initialization stage, $N_{an+1}$ represents the value updated in the nth stage. $\phi$ is a loss function used to optimize ambient light estimation A.
gradient $\phi$  Represents the loss function of optimized ambient light estimation $\phi$  The gradient is used to update the value of A.

So, the entire iterative process mainly optimizes the shadow removal effect in the input image Is by updating the ambient light estimation A and the shadow removal image Ins. Advanced technologies such as U-net architecture, ViT, and AIGC Models are applied in neural network design to improve model performance.

We have designed a basic convolutional module called Dynamic Mapping Residual Block (DMRB) for each scale of the network. And use two CNN networks to calculate the initial and gradient results respectively.
Formalize this problem as a maximum posterior estimation problem. We transform the shadow removal problem into solving the minimum variational model, and use it to further transform it into an optimization problem without constrained optimization:
\begin{equation}
    \arg\min\limits_{Ins}D(Is,Ins,A)+\beta g(A)+\lambda \phi(A)
\end{equation}

Among them, $\beta$ is a weight parameter. Since the data fidelity terms $D (\cdot)$ and $g (\cdot)$ are quadratic constraints, they are differentiable. If we assume $\lambda'(\cdot)$ is also differentiable, and in each iteration step, we need to alternately update A and Ins as follows:
\begin{equation}
\left\{\begin{array}{c}
A^{n+1}=A^n-\eta_A\left(\nabla_A D^n+\beta \nabla_A g\left(A^n\right)+\lambda \nabla_A \phi\left(A^n\right)\right) \\
I_{n s}^{\mathrm{n}+1}=\left(1+A^{n+1}\right) \star I s,
\end{array}\right.
\end{equation}

Among them, $\nabla$ represents the gradient operator; n represents the current number of iterations, with a maximum of K.

During the above iteration process, equation (14) has two unresolved problems. The first problem is to calculate the gradient of the objective function. The corresponding gradients of quadratic terms $D (\cdot)$ and $g (\cdot)$ are clearly easy to calculate. Then, according to Liu et al. (2019), the corresponding $\lambda'(\cdot)$ can be achieved through deep CNN A. $\nabla A \lambda'(An)$ can be directly learned from the training dataset. Another issue is how to obtain the initial values of the iterative algorithm. Due to the ability of CNN to produce reasonable results, we use CNN technology to estimate the initial values A0 and Is.

\subsection{Data fidelity and regularization Network Loss}

Our network loss consists of two parts: the data fidelity term and the regularization term of map A. Specifically, we use the mean square error (MSE) as the loss function, which is defined as:
\begin{equation}
L=\sum_{n=0}^k Y\left|I-I n s_n\right|^2+\sum_{n=0}^k \gamma{ }_g\left|(1-M) \times A_n\right|^2
\end{equation}

Where I and $A_n$ represents the estimated shadow removal results and learned transformation maps for the $n^{th}$ iteration stage, respectively; $\gamma$ and $\gamma{ }_g$ represents the corresponding trade-off weight.

\subsection{Training Details}

We train the network in each stage end-to-end, and use the Adam optimizer to optimize the loss function. The training set includes a large amount of synthesized image data and real-world shadow images. At each stage, we use advanced inpainting technologies such as SOTA Interpreters to repair the shadow removed image, and use AIGC Models to generate the required content in the hole.

\section{Experiment}
\subsection{Dataset and performance comparisons with existing methods}
This paper uses the latter dataset ISTD \cite{2}, SRD.  We chose the Adam optimizer with a $\beta$ of 0.999 at the same time.

%\begin{figure}[h]
%\centering
%\includegraphics[width=3in]{result.png}
%\caption{Results of SpT-former(The first column on the left is the input image, the second column is the de-shadowing result, the third column is the attentional diagram, and the fourth column is the original image)}
%\label{result}
%\end{figure}
Our method is compared with existing methods including Yang \cite{17}, Guo \cite{15}, Gong \cite{16}, DeShadowNet\cite{1}, STC-GAN \cite{2}, DSC \cite{4}, Mask-ShadowGAN \cite{9}, RIS-GAN \cite{5}, DHAN \cite{6}, SID \cite{3}, LG-shadow \cite{10}, G2R \cite{11}, SG-ShadNet \cite{SG-ShadowNet}, Auto-exp \cite{Auto-exp}, Bejective \cite{24}, SpA-Former \cite{SpA-Former}. We adopt RMSE, SSIM and PSNR in the LAB color space as evaluation metrics.
\par
We evaluate the performance of different methods on the shadow regions, non-shadow regions, and the whole image. We can see that the proposed SAM-helps-shadow achieves the satisfactory performance among all the compared methods as shown in Fig.\ref{compare}. We also report the shadow removal performance of our proposed method on the ISTD and srd dataset. As shown in Table.\ref{compare-table-itsd} and Table.\ref{compare-table-srd}. According to the Table, our suggested SpA-Former gets the approving RMSE values in shadow areas, non-shadow regions, and the complete image on the ISTD. This implies that the recovered shade-removal pictures generated by our SAM-helps-Shaodw are substantially closer to the corresponding ground-truth shadow-free images.
\par
%\textbf{\textsl{By the way, the current image de-shadowing networks are roughly divided into three categories, weakly supervised, supervised, and unsupervised. Regardless of which one, the current common method is the two-stage method, which first detects and then removes the shadow part, however, the training time and GPU performance required for the two-stage method are relatively high. Our proposed SpA-former belongs to the one-stage network, and he is able to remove the shadow part while recognizing it through the spatial attention map, and more importantly, the network in this paper can be trained and used on an ordinary 1080Ti, which greatly reduces the application barriers.}}

\begin{table*}\caption{Performance comparison of results on ISTD (RMSE-N/RMSE-S represent no shadow and shadow espectively).}
\centering
\resizebox{14cm}{2.4cm}{
%\scalebox{0.8}{
\begin{tabular}{c|c|c|c|c|c|c|c|c|c}

%\hline & \multicolumn{10}{c} {\text {ISTD}} \\
\hline
\text{Model}  & \text { RMSE-all}$\downarrow$ & \text {RMSE-N}$\downarrow$ & \text{RMSE-S}$\downarrow$ & \text{SSIM}$\uparrow$  & \text{SSIM-N}$\uparrow$ & \text{SSIM-S}$\uparrow$ & \text{PSNR}$\uparrow$  & \text{PSNR-N}$\uparrow$ & \text{PSNR-S}$\uparrow$\\
\hline
\text {Yang [TIP2012]\cite{17}}   & \ 15.63   &  14.83  & 19.82 & -  & - & - & - & - & -    \\
\hline \text {Guo [TPAMI2013]\cite{15}}    &   9.3     & 7.46    & 18.95 &0.919 &0.944 &0.978 &23.07 &24.86 &30.98 \\
\hline \text { Gong [BMVC2014]\cite{16} } & 8.53      & 7.29    & 14.98 &0.908 &0.929 &0.98 &24.07 &25.26 &32.43\\
\hline \text { DeShadowNet[CVPR2017]\cite{1} } & 7.83 & 7.19 & 12.76 & -  & - & - & - & - & -  \\
\hline \text { STC-GAN [CVPR2018]\cite{2} }     & 7.47 & 6.93 &10.33 &0.929 &0.947 &0.985 &27.43 &28.67 &35.8\\
\hline \text { SID [ICCV2019]\cite{3} }             &7.96  &7.72	&9.64	&0.948	&0.964	&0.986	&25.01	&26.1	&32.88\\
\hline \text {DSC [TPAMI2019]\cite{4}}               & 6.67 & 6.39 & 9.22 &0.845 &0.885 &0.967 &26.62 &28.18 &33.45\\
\hline\text {RIS-GAN [AAAI2019]\cite{5}}            &6.62  &6.31	 &9.15 &-  & - & - & - & - & - \\
\hline \text {Mask-ShadowGAN[2019]\cite{9}}    &7.63  &7.03	 &10.35 &-  & - & - & - & - & -  \\
\hline\text {DHAN [AAAI2020]\cite{6}}               &6.28  &5.92	&8.43	&0.921	&0.941	&0.983	&27.88	&29.54	&34.79\\
\hline\text {LG-shadow [ECCV2020]\cite{10}}      &6.67  &5.93	&11.51	&0.906	&0.938	&0.974	&25.83	&28.32	&31.08\\
\hline\text {Auto-Exp [CVPR2021]\cite{Auto-exp} }    &5.88 &5.51	&7.9	&0.845	&0.879	&0.975	&27.19	&28.6	&34.71\\
\hline\text {G2R [CVPR2021]\cite{11}}          &7.84  &7.54  &10.71 &0.932 &0.967	&0.974	&24.72	&26.18	&31.62\\
\hline\text { Bijetive [CVPR2022]\cite{24}}        &\textbf{4.78}  & \textbf{4.28}	&\textbf{6.76}	&0.928	& 0.948	&0.983	&\textbf{29.02}	&31.14 &34.84 \\
\hline\text { SpA-Former [IJCNN2023]\cite{SpA-Former}}        &6.86  & 6.22	&10.48	&0.931	& 0.956	&0.982	&27.73	&30.16&33.51 \\
\hline \text {SAM-helps-shadow}                                     &5.09	&4.55	&8.29	&\textbf{0.959}	&\textbf{0.978}& \textbf{0.986} &28.85	&\textbf{31.54} &\textbf{36.95}\\
\hline
\end{tabular}}
\label{compare-table-itsd}
\end{table*}

\begin{figure}[h]
\centering
\includegraphics[width=3.5in]{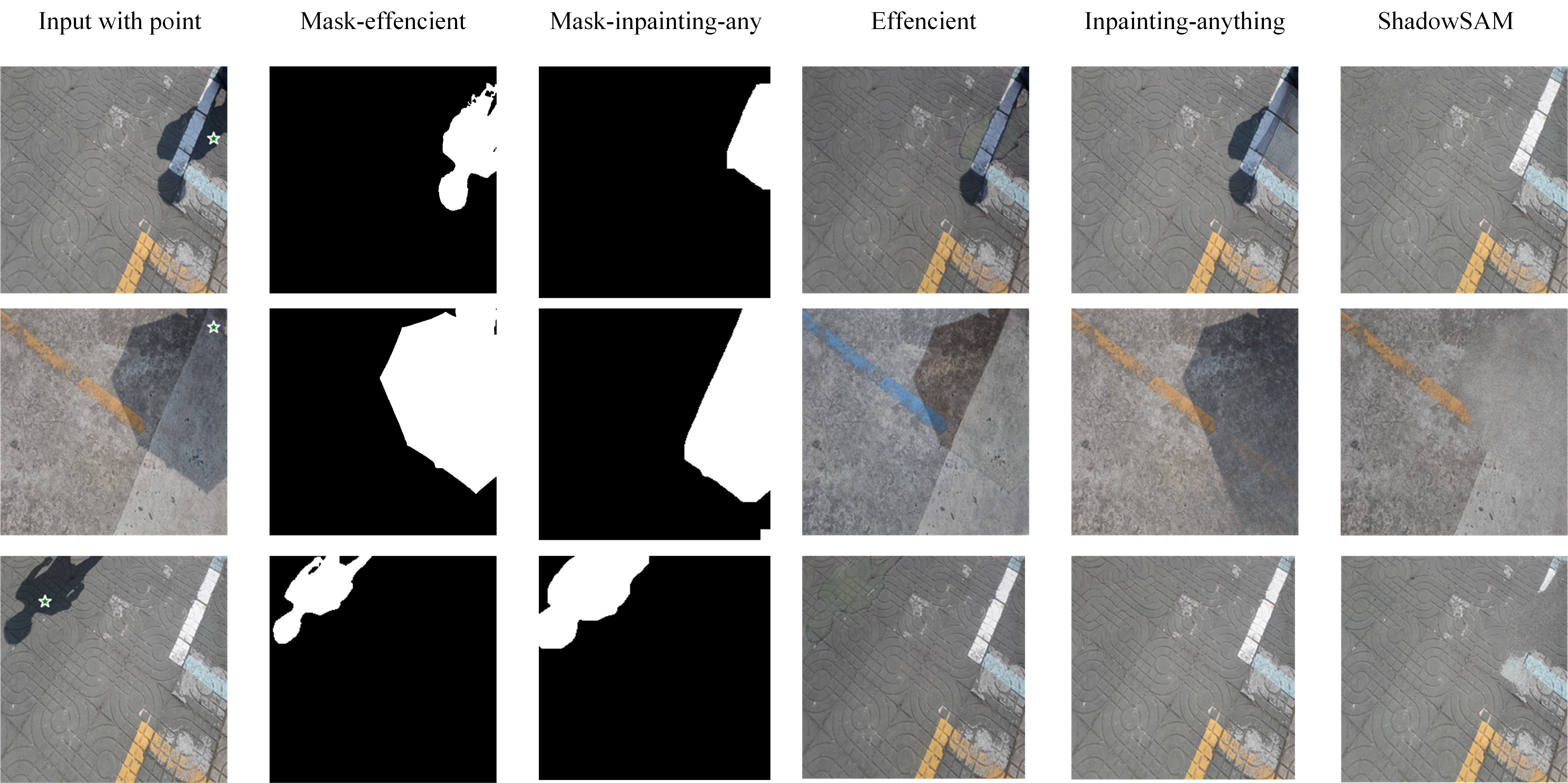}
\caption{The compare results of Shadow-helps-shadow and other methods}
\label{ablation}
\end{figure}

\begin{table*}[t]\caption{Performance comparison of results on SRD (RMSE-N/RMSE-S represent no shadow and shadow espectively).}
\centering
\resizebox{14cm}{1.5cm}{
\begin{tabular}{c|c|c|c|c|c|c|c|c|c}

%\hline & \multicolumn{10}{c} {\text {ISTD}} \\
\hline
\text{Model}  & \text { RMSE-all}$\downarrow$ & \text {RMSE-N}$\downarrow$ & \text{RMSE-S}$\downarrow$ & \text{SSIM}$\uparrow$  & \text{SSIM-N}$\uparrow$ & \text{SSIM-S}$\uparrow$ & \text{PSNR}$\uparrow$  & \text{PSNR-N}$\uparrow$ & \text{PSNR-S}$\uparrow$\\
\hline \text {Guo[TPAMI2013]\cite{15}}          &14.05   & 6.47 & 29.89 & - & - & - &- &- &- \\
\hline \text { DeShadowNet[CVPR2017]\cite{1} } & 6.64 & 4.84 & 11.78 & -  & - & - & - & - & -  \\
\hline \text {DSC[TPAMI2019]\cite{4}}          & 17.43&  16.45& 20.09 &0.648 &0.760 &0.905 &20.65 &23.35 &25.83\\
\hline\text {DHAN[AAAI2020]\cite{6}}      &4.80 &3.77	&6.76	&0.928	&0.947	&0.972 &28.93&32.49	&32.63\\
\hline DC-ShadowNet[CVPR2021]\cite{DC-shadowNet}  & 6.15&5.18 & 8.96 & 0.864  & 0.927 & 0.957 & 27.08& 30.04 & 31.33 \\
\hline\text {Auto-Exp[CVPR2021]\cite{Auto-exp}}       &6.75 &5.97 &9.02	&0.856	&0.920	&0.956	&26.72	&29.37	&31.43\\
\hline\text { Bejetive[CVPR2022]\cite{24}}        &\textbf{4.15}  & 3.21	&\textbf{5.92}	&0.944	& 0.977	&0.977	&30.35	&34.31 &33.91 \\
\hline\text { SpA-Former[IJCNN2023]\cite{SpA-Former}}        &6.86  & 6.22	&10.48	&0.931	& 0.956	&\textbf{0.982}	&27.73	&30.16&33.51 \\
\hline \text {SAM-helps-shadow}           &4.79	&\textbf{3.74}	&7.44	&\textbf{0.952}	&\textbf{0.981}	&0.979	&30.72 & \textbf{33.85} &\textbf{33.94}\\
\hline
\end{tabular}}

\label{compare-table-srd}
\end{table*}

%\begin{table}[H]\setlength{\tabcolsep}{4mm}
%\caption{FLOPs and Params comparisons.}
%\begin{tabular}{cccc}
%%\hline & \multicolumn{10}{c} {\text {ISTD}} \\
%\toprule
%           \textbf{Method}                     &  \textbf{FLOPs}           & \textbf{Total Params}     & \textbf{Training Time}
%\\
%\midrule    DGUNet \cite{DGUNet}                 	  &$8.4278\times 10^{11}$	 &12,176,119        & 94,600 s
%\\
%\midrule    MPRNet \cite{MPRNet}                     	  &$2.1946\times 10^{12}$	 &3,637,249         & 41,400 s
%\\
%\midrule    UIR-Net           	  &$2.4120\times 10^{11}$   &3,632,740         & 22,000 s
%\\
%\bottomrule
%\end{tabular}
%\label{Flops-and-params}
%\end{table}

\section{Ablation Study on SAM-helps-Shadow}
The Inpainting-anything \cite{inpainting-anything} and Efficient \cite{24} technique was first examined. Despite its popularity, our results indicate that this method’s performance was subpar in terms of image segmentation. The Image Inpainting method was prone to misinterpret the segmentation areas and had challenges providing clear delineation.

On comparing these methods, we discovered that our novel segmentation technique outperformed both the Image Inpainting and SAM methods in terms of image segmentation and de-shadowing. Our method delivered a more precise segmentation with fewer erroneous delineations. Furthermore, the de-shadowing was more effective, resulting in clearer images with better details.

While both the Image Inpainting and SAM techniques have their merits and can provide satisfactory results under certain conditions, our comparative ablation study demonstrated that our proposed technique could offer better performance in image segmentation and de-shadowing tasks. Future work will aim to enhance the robustness and versatility of our proposed method, as well as expanding its application to a broader range of scenarios.

\section{Results}
The challenges surrounding the application of image shadow removal to real-world images and not just constrained datasets like ISTD/SRD have highlighted an urgent need for zero-shot learning in this field. In this study, we innovatively adapted the SAM (Segment anything model) for shadow removal by introducing SAM-helps-Shadow, effectively integrating shadow detection and removal into a single stage. Our approach utilized the model's detection results as a potent prior for facilitating shadow detection, followed by shadow removal using a second-order deep unfolding network

\par


\begin{thebibliography}{99}

\bibitem{1}Qu L, Tian J, He S, et al. Deshadownet: A multi-context embedding deep network for shadow removal//Proceedings of the IEEE Conference on Computer Vision and Pattern Recognition. 2017: 4067-4075.
\bibitem{2}Wang J, Li X, Yang J. Stacked conditional generative adversarial networks for jointly learning shadow detection and shadow removal//Proceedings of the IEEE Conference on Computer Vision and Pattern Recognition. 2018: 1788-1797.
\bibitem{3}Le H, Samaras D. Shadow removal via shadow image decomposition//Proceedings of the IEEE/CVF International Conference on Computer Vision. 2019: 8578-8587.
\bibitem{4}Hu X, Fu C W, Zhu L, et al. Direction-aware spatial context features for shadow detection and removal. IEEE transactions on pattern analysis and machine intelligence, 2019, 42(11): 2795-2808.
\bibitem{5}Zhang L, Long C, Zhang X, et al. Ris-gan: Explore residual and illumination with generative adversarial networks for shadow removal//Proceedings of the AAAI Conference on Artificial Intelligence. 2020, 34(07): 12829-12836.
\bibitem{6}Cun X, Pun C M, Shi C. Towards ghost-free shadow removal via dual hierarchical aggregation network and shadow matting GAN//Proceedings of the AAAI Conference on Artificial Intelligence. 2020, 34(07): 10680-10687.
\bibitem{7}Fu L, Zhou C, Guo Q, et al. Auto-exposure fusion for single-image shadow removal//Proceedings of the IEEE/CVF Conference on Computer Vision and Pattern Recognition. 2021: 10571-10580.
\bibitem{8}Chen Z, Long C, Zhang L, et al. CANet: A Context-Aware Network for Shadow Removal//Proceedings of the IEEE/CVF International Conference on Computer Vision. 2021: 4743-4752.
\bibitem{9}Hu X, Jiang Y, Fu C W, et al. Mask-ShadowGAN: Learning to remove shadows from unpaired data//Proceedings of the IEEE/CVF International Conference on Computer Vision. 2019: 2472-2481.
\bibitem{10}Liu Z, Yin H, Mi Y, et al. Shadow removal by a lightness-guided network with training on unpaired data. IEEE Transactions on Image Processing, 2021, 30: 1853-1865.
\bibitem{11}Vasluianu F A, Romero A, Van Gool L, et al. Self-Supervised Shadow Removal. arXiv preprint arXiv:2010.11619, 2020.
\bibitem{SG-ShadowNet}Wan J, Yin H, Wu Z, et al. Style-Guided Shadow Removal//European Conference on Computer Vision. Springer, Cham, 2022: 361-378.

\bibitem{Auto-exp}Fu L, Zhou C, Guo Q, et al. Auto-exposure fusion for single-image shadow removal//Proceedings of the IEEE/CVF conference on computer vision and pattern recognition. 2021: 10571-10580.
\bibitem{UIR-Net}Mei X, Ye X, Zhang X, et al. UIR-Net: A Simple and Effective Baseline for Underwater Image Restoration and Enhancement. Remote Sensing, 2022, 15(1): 39
\bibitem{LGNet}Qiu L, Yu D, Zhang C, et al. A Local¨CGlobal Framework for Semantic Segmentation of Multisource Remote Sensing Images[J]. Remote Sensing, 2022, 15(1): 231.
\bibitem{12}Le H, Samaras D. From shadow segmentation to shadow removal//European Conference on Computer Vision. Springer, Cham, 2020: 264-281.

\bibitem{TC-GAN}Tan C, Feng X. Unsupervised Shadow Removal Using Target Consistency Generative Adversarial Network. arXiv preprint arXiv:2010.01291, 2020.
\bibitem{13}Liu Z, Yin H, Wu X, et al. From shadow generation to shadow removal//Proceedings of the IEEE/CVF Conference on Computer Vision and Pattern Recognition. 2021: 4927-4936.

\bibitem{14}Tan C, Feng X. Unsupervised Shadow Removal Using Target Consistency Generative Adversarial Network. arXiv preprint arXiv:2010.01291, 2020.
\bibitem{22}He Y, Xing Y, Zhang T, et al. Unsupervised Portrait Shadow Removal via Generative Priors//Proceedings of the 29th ACM International Conference on Multimedia. 2021: 236-244.
\bibitem{23}Zhu Y, Xiao Z, Fang Y, et al. Efficient Model-Driven Network for Shadow Removal. 2022.
\bibitem{24}Zhu Y, Huang J, Fu X, et al. Bijective Mapping Network for Shadow Removal//Proceedings of the IEEE/CVF Conference on Computer Vision and Pattern Recognition. 2022: 5627-5636.
\bibitem{15}R. Guo, Q. Dai, and D. Hoiem, Paired regions for shadow detection and removal, IEEE TPAMI, vol. 35, no. 12, 2012.
\bibitem{SiENet}Zhang X, Chen F, Wang C, et al. Sienet: Siamese expansion network for image extrapolation. IEEE Signal Processing Letters, 2020, 27: 1590-1594.
\bibitem{16}H. Gong and D. Cosker, Interactive shadow removal and ground truth for variable scene categories, in Proc. BMVC, 2014.
\bibitem{17}Q. Yang, K. Tan, and N. Ahuja. Shadow removal using bilateral filtering. IEEE TIP, 21(10):4361??C4368, 2012.
\bibitem{18}Pan H. Cloud removal for remote sensing imagery via spatial attention generative adversarial network. arXiv preprint arXiv:2009.13015, 2020.
\bibitem{IJPRAI}Shen R, Zhang X, Xiang Y. AFFNet: attention mechanism network based on fusion feature for image cloud removal. International Journal of Pattern Recognition and Artificial Intelligence, 2022: 2254014.
\bibitem{unzi2}Yu Q, Zheng N, Huang J, et al. CNSNet: A Cleanness-Navigated-Shadow Network for Shadow Removal. arXiv preprint arXiv:2209.02174, 2022.
\bibitem{unzi3}Jin Y, Yang W, Ye W, et al. ShadowDiffusion: Diffusion-based Shadow Removal using Classifier-driven Attention and Structure Preservation[J]. arXiv preprint arXiv:2211.08089, 2022.
\bibitem{unzi4}Wan J, Yin H, Wu Z, et al. CRFormer: A Cross-Region Transformer for Shadow Removal. arXiv preprint arXiv:2207.01600, 2022.
\bibitem{unzi6}Xu Y, Lin M, Yang H, et al. Shadow-Aware Dynamic Convolution for Shadow Removal. arXiv preprint arXiv:2205.04908, 2022.
\bibitem{zhu2020}Zhu T, Xia S, Bian Z, Lu C. Highlight removal in facial images. InPattern Recognition and Computer Vision: Third Chinese Conference, PRCV 2020, Nanjing, China, October 16¨C18, 2020, Proceedings, Part I 3 2020 (pp. 422-433). Springer International Publishing.
\bibitem{wu2021}Wu X, Lu C, Gu C, Wu K, Zhu S. Domain adaptation for viewpoint estimation with image generation. In2021 International Conference on control, automation and information sciences (ICCAIS) 2021 Oct 14 (pp. 341-346). IEEE.
\bibitem{lu2018}Lu C, Wang H, Gu C, Wu K, Guan X. Viewpoint estimation for workpieces with deep transfer learning from cold to hot. InNeural Information Processing: 25th International Conference, ICONIP 2018, Siem Reap, Cambodia, December 13-16, 2018, Proceedings, Part I 25 2018 (pp. 21-32). Springer International Publishing.
\bibitem{lu2020}Lu C, Gu C, Wu K, Xia S, Wang H, Guan X. Deep transfer neural network using hybrid representations of domain discrepancy. Neurocomputing. 2020 Oct 7;409:60-73.
\bibitem{lu2023}Lu C, Zhu H, Koniusz P. From Saliency to DINO: Saliency-guided Vision Transformer for Few-shot Keypoint Detection. arXiv preprint arXiv:2304.03140. 2023 Apr 6.
\bibitem{Jiang2022}Jiang Y, Yang F, Bian Z, Lu C, Xia S. Mask removal: Face inpainting via attributes. Multimedia Tools and Applications. 2022 Sep;81(21):29785-97.
\bibitem{lu2019}Lu C, Xia S, Shao M, Fu Y. Arc-support line segments revisited: An efficient high-quality ellipse detection. IEEE Transactions on Image Processing. 2019 Aug 15;29:768-81.
\bibitem{lu2017}Lu C, Xia S, Huang W, Shao M, Fu Y. Circle detection by arc-support line segments. In2017 IEEE International Conference on Image Processing (ICIP) 2017 Sep 17 (pp. 76-80). IEEE.
\bibitem{inpainting-anything}Yu T, Feng R, Feng R, et al. Inpaint anything: Segment anything meets image inpainting[J]. arXiv preprint arXiv:2304.06790, 2023.

%%%%%%SAM
\bibitem{SAM}Kirillov A, Mintun E, Ravi N, et al. Segment anything[J]. arXiv preprint arXiv:2304.02643, 2023.
\bibitem{medsam}Ma J, Wang B. Segment anything in medical images[J]. arXiv preprint arXiv:2304.12306, 2023.
\bibitem{cmedsam}Zhang K, Liu D. Customized segment anything model for medical image segmentation[J]. arXiv preprint arXiv:2304.13785, 2023.
\bibitem{sammd}Roy S, Wald T, Koehler G, et al. Sam. md: Zero-shot medical image segmentation capabilities of the segment anything model[J]. arXiv preprint arXiv:2304.05396, 2023.
\bibitem{rssam}Wang D, Zhang J, Du B, et al. Scaling-up Remote Sensing Segmentation Dataset with Segment Anything Model[J]. arXiv preprint arXiv:2305.02034, 2023.
\bibitem{samadpter}Chen T, Zhu L, Ding C, et al. SAM Fails to Segment Anything?--SAM-Adapter: Adapting SAM in Underperformed Scenes: Camouflage, Shadow, and More[J]. arXiv preprint arXiv:2304.09148, 2023.
\bibitem{samdomain}Jing Y, Wang X, Tao D. Segment anything in non-euclidean domains: Challenges and opportunities[J]. arXiv preprint arXiv:2304.11595, 2023.
\bibitem{SpA-Former}Zhang X F, Gu C C, Zhu S Y. SpA-Former: Transformer image shadow detection and removal via spatial attention[J]. arXiv e-prints, 2022: arXiv: 2206.10910.
\bibitem{DC-shadowNet}Jin Y, Sharma A, Tan R T. DC-ShadowNet: Single-Image Hard and Soft Shadow Removal Using Unsupervised Domain-Classifier Guided Network[C]//Proceedings of the IEEE/CVF International Conference on Computer Vision. 2021: 5027-5036.
\end{thebibliography}
\end{document}